\documentclass[11pt,a4paper]{article}
\usepackage{times,latexsym}
\usepackage{url}
\usepackage[T1]{fontenc}

\usepackage{amsmath}
\usepackage{graphicx}
\usepackage{subfigure} 
\usepackage{hyperref} 
\usepackage{cleveref} 

\usepackage{algorithm,algorithmic} 
\usepackage{tcolorbox}
\usepackage{booktabs}
\usepackage{multirow} 
\usepackage{amssymb}
\usepackage{enumitem}

\usepackage[acceptedWithA]{tacl2021v1}

\usepackage{xspace,mfirstuc,tabulary}

\newif\iftaclinstructions
\taclinstructionsfalse 
\iftaclinstructions
\renewcommand{\confidential}{}
\renewcommand{\anonsubtext}{(No author info supplied here, for consistency with
TACL-submission anonymization requirements)}
\newcommand{\instr}
\fi

\iftaclpubformat 

\else

\fi

\newcolumntype{L}[1]{>{\raggedright\arraybackslash}p{#1}}
\newcolumntype{C}[1]{>{\centering\arraybackslash}p{#1}}
\newcolumntype{R}[1]{>{\raggedleft\arraybackslash}p{#1}}

\DeclareSymbolFont{extraup}{U}{zavm}{m}{n}
\DeclareMathSymbol{\varheart}{\mathalpha}{extraup}{86}
\DeclareMathSymbol{\vardiamond}{\mathalpha}{extraup}{87}

\usepackage{amsmath}
\usepackage{amssymb}
\usepackage{wasysym}

\DeclareSymbolFont{extraup}{U}{zavm}{m}{n}
\DeclareMathSymbol{\varspadesuit}{\mathalpha}{extraup}{83}
\DeclareMathSymbol{\varheartsuit}{\mathalpha}{extraup}{86}
\DeclareMathSymbol{\vardiamond}{\mathalpha}{extraup}{87}
\DeclareMathSymbol{\varclubsuit}{\mathalpha}{extraup}{88}

\title{{Fusing Large Language Models with\\Temporal Transformers for Time Series Forecasting}}

\author{
    Chen Su$^{{\spadesuit}}$, \hspace{0.1cm}
    Yuanhe Tian$^{\varheart}$, \hspace{0.1cm}
    Qinyu Liu$^{\varclubsuit}$, \hspace{0.1cm}
    Jun Zhang$^{\Diamond}$, \hspace{0.1cm}
    Yan Song$^{{\spadesuit}*}$
    \\
    $^{\spadesuit}$University of Science and Technology of China \hspace{0.1cm}
    $^{\varheart}$University of Washington \\
    $^{\varclubsuit}$Beijing Northern Computility InterConnection Co., Ltd. \hspace{0.1cm}
    $^{\Diamond}$ENN Group Co., Ltd. \\
    $^{\spadesuit}$\texttt{suchen4565@mail.ustc.edu.cn} \hspace{0.1cm}
    $^{\varheart}$\texttt{yhtian@uw.edu} \\  
    $^{\varclubsuit}$\texttt{garryliuqy@hotmail.com} \hspace{0.1cm}
    $^{\Diamond}$\texttt{zhangjunbp@enn.cn} \hspace{0.1cm}
    $^{\spadesuit}$\texttt{clksong@gmail.com}
}

\date{}

\begin{document}

\maketitle

\renewcommand{\thefootnote}{\fnsymbol{footnote}}
\footnotetext[1]{Corresponding author.}
\renewcommand{\thefootnote}{\arabic{footnote}}

\begin{abstract}
{
Recently, large language models (LLMs) have demonstrated powerful capabilities in performing various tasks and thus are applied by recent studies to time series forecasting (TSF) tasks, which predict future values with the given historical time series.
Existing LLM-based approaches transfer knowledge learned from text data to time series prediction using prompting or fine-tuning strategies.
However, LLMs are proficient at reasoning over discrete tokens and semantic patterns but are not initially designed to model continuous numerical time series data.
The gaps between text and time series data lead LLMs to achieve inferior performance to a vanilla Transformer model that is directly trained on TSF data.
However, the vanilla Transformers often struggle to learn high-level semantic patterns.
In this paper, we design a novel Transformer-based architecture that complementarily leverages LLMs and vanilla Transformers, so as to integrate the high-level semantic representations learned by LLMs into the temporal information encoded by time series Transformers, where a hybrid representation is obtained by fusing the representations from the LLM and the Transformer.
The resulting fused representation contains both historical temporal dynamics and semantic variation patterns, allowing our model to predict more accurate future values.
Experiments on benchmark datasets demonstrate the {effectiveness of the proposed approach.}\footnote{The code is available at \url{https://github.com/synlp/SemInf-TSF}.}
}
\end{abstract}

\section{Introduction}
\label{introduction}

Time series forecasting (TSF) is a task that focuses on predicting future values of sequential data points based on historical observations,
which aims to capture latent temporal patterns and dependencies within dynamic systems to extrapolate future fluctuations.
This task holds significant practical value, enabling informed decision-making across diverse domains such as finance for stock prediction \cite{tsf4stock1, tsf4stock2, tsf4stock3}, meteorology for weather modeling \cite{tsf4weather1, tsf4weather2, tsf4weather3}, energy management for load forecasting \cite{tsf4energy1, tsf4energy2, tsf4energy3}, and healthcare for epidemic trend analysis \cite{tsf4epidemic1, tsf4epidemic2, tsf4epidemic3}.

\begin{figure*}[t]
    \centering
    \includegraphics[width=\linewidth, trim=0 50 0 10]{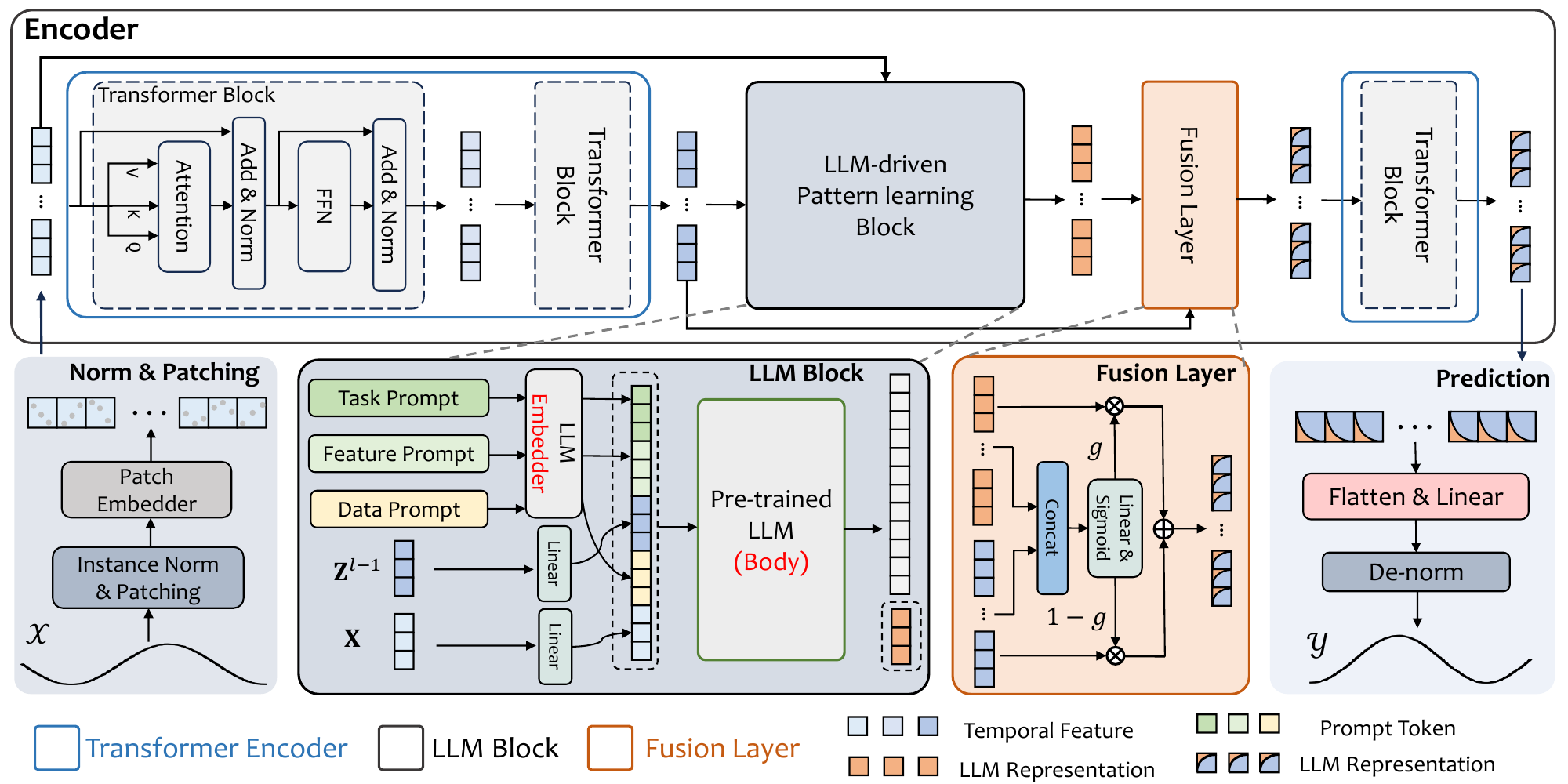}
    \caption{The overall architecture of our approach.}
    \label{fig:model}
    \vspace{-0.2cm}
\end{figure*}

Early TSF approaches \cite{TSRnn2, TSRnn1, TSConv1, TSConv2, TSConv3, TSRnn3} relying on recurrent and convolutional architectures exhibit poor long-range modeling capabilities, leading to suboptimal performance in long-term predictions.
Transformer-based approaches \cite{informer, autoformer, fedformer} address such challenges by leveraging self-attention mechanisms for global temporal relationships, yet their effectiveness remains constrained by insufficient training data and model scale.
Recent breakthroughs of large language models (LLMs) demonstrate that billions of parameters and pre-training on large corpora are effective in acquiring universal knowledge that allows them to perform well in downstream NLP tasks through prompt engineering \cite{NLPprompt1, NLPprompt3, NLPprompt2} and fine-tuning \cite{NLPfinetune2,NLPfinetune3,NLPfinetune4}.
This observation inspires studies \cite{time-llm, llmtime, gpt4mts} to apply LLMs to TSF, with, however, critical challenges persist in adapting LLMs to process continuous sequential data and interpret forecasting task objectives.
In general,
LLM-based approaches for TSF are primarily categorized into two main streams.
{The first category transforms time series into text, thereby enabling LLMs to process them as textual data.}
The second integrates trainable embedding layers that convert time series data into continuous token representations, thus preserving full temporal information. 
{
The approaches in both streams leverage LLM as time series ``encoders'', using the output of the last LLM layer to predict time series values.
However, these approaches face several issues.
First, they apply LLMs pre-trained on textual data to encode time series.
The LLMs carry a strong prior bias towards the text domain and thus may misinterpret the input time series.
For instance, LLMs tend to treat time series embeddings as a special type of tokens, and these time series tokens within the LLM's embedding space may have high similarity to word embeddings that are irrelevant to time series understanding.
While some studies \cite{timecma, calf} introduce modal alignment between time-series and synthetic text to mitigate this conflict, such synthetic text often reflects human subjectivity and typically fails to achieve true matching.
Second, these approaches utilize LLMs with a decoder-only architecture for time-series encoding. 
This inherent decoder-only structure restricts the information integrity during encoding.
Consequently, the representations output by LLMs for time-series often reflect a unidirectional perspective.
Therefore, directly employing LLMs for TSF is challenging and needs further accommodations.
Considering that conventional deep learning approaches (e.g., a vanilla Transformer) are able to effectively learn features in time series data from scratch, it is expected to combine LLMs and the conventional approaches to better perform TSF.
}

{
In this paper, we propose a novel TSF model that benefits from both LLM and vanilla Transformer.
It aims to fuse the semantic patterns computed by LLMs from time series data into the representations generated by a Transformer-based time series encoder, achieving a synergistic integration of temporal dynamics and high-level semantic information.
Specifically, the model employs a vanilla Transformer encoder to process input time series patches, producing representations containing temporal dynamics.
Subsequently, the temporal representations and the original time series patches are fed into an LLM.
The LLM identifies semantic patterns among the time series tokens and leverages them accordingly to generate a higher-level semantic representation. 
A gate mechanism then performs feature fusion on these two distinct representation types, i.e., the temporal dynamics from the Transformer and the semantic patterns from the LLM, resulting in a comprehensive and integrated representation.
This fused representation is then used to generate future time series values.
Extensive experiments on benchmark datasets demonstrate that our framework surpasses existing TSF approaches, achieving state-of-the-art results across diverse application domains.
}

\section{The Approach}

{
Given a historical time series $\mathcal{X}$ with the historical length \( T_x \), the objective of TSF is to forecast future values $\mathcal{Y}$ over \( T_y \) time steps.
Our approach for TSF is illustrated in Figure \ref{fig:model} with three core components: a Transformer-based backbone model that encodes the time series $\mathcal{X}$ and forecasts the future values $\mathcal{Y}$, an LLM that generates semantic representations of the time series, and a feature fusion module incorporates the features from the LLM into the backbone Transformer.
Specifically, the time series $\mathcal{X}$ is divided into patches, and the patches are fed into the Transformer-based backbone with $L$ layers of multi-head attentions (MHA) and a linear project layer that takes the output of the last HMA layer and forecast the future values $\mathcal{Y}$.
The output of the ($l$-1)-th MHA layer (denoted as $\mathbf{Z}^{l-1}$), the original time series $\mathcal{X}$, and prompts are fed into the LLM to compute the deep semantics of $\mathcal{X}$, which is denoted as $\mathbf{X}_s$.
The feature fusion module then integrates the output of the LLM $\mathbf{X}_s$ into the $l$-th MHA layer of the backbone Transformer for further processing.
In the following text, we firstly present the Transformer-based backbone model for TSF, then illustrate the process of prompting the LLM to encode the time series data, and finally the feature fusion module that integrates the LLM output into the backbone model. 
}

\subsection{Transformer-based Backbone Model}

{
The Transformer-based backbone model consists of a Transformer encoder with $L$ layers of MHA and a fully connected layer $f_\text{F}$.
The Transformer encoder aims to capture temporal dependencies and encode sequential information from the input time series. 
This input time series $\mathcal{X}$ undergoes patching with a stride $S$ and a patch length $T$, dividing it into a sequence of overlapping patches $\mathbf{X} \in \mathbb{R}^{N \times T}$. 
The patch sequence length $N$ is determined by
\begin{equation}
    N = \left\lfloor \frac{(T_x - T)}{S} \right\rfloor + 2
\end{equation}
where $\left\lfloor \cdot \right\rfloor$ means the floor function that rounds a number down to the nearest integer.
Especially, we pad the input time series at the end by repeating the last value $S$ times to ensure that the values of the last part form a patch.
This patching reduces sequence length and facilitates the capture of local patterns. 
A linear projection layer with a trainable parameter $\mathbf{W}_t$ transforms the input patches into a high-dimensional feature space, where the positional encoding $\mathbf{E}_\text{pos}$ is added to obtain the input embedding (i.e., denoted as $\mathbf{Z}$) of the Transformer. 
This entire process is expressed by:
\begin{equation}
    \mathbf{Z} = \mathbf{X} \mathbf{W}_t + \mathbf{E}_\text{pos}
\end{equation}
The resulting sequence $\mathbf{Z}$ serves as input to the $L$-layer MHA, where the output of a Transformer layer is the input of the next layer.
Particularly, at the layer $l$, we enhance its input (denoted as $\mathbf{Z}^{l-1}$) with the output of the LLM (see the following subsections for details), so as to better encode the time series data.
The output of the last MHA layer, i.e., $\mathbf{Z}^{L}$, is flattened and fed into the fully connected projection layer for forecasting the future time series $\widehat{\mathcal{Y}}$.
The entire model minimizes the mean squared error (MSE) loss $\mathcal{L}$ between the predicted time series $\widehat{\mathcal{Y}}$ and the ground truth time series $\mathcal{Y}$:
\begin{equation}
    \mathcal{L} = \frac{1}{T_y} \sum_{i=1}^{T_y} \left( \widehat{\mathcal{Y}}_i - \mathcal{Y}^{*}_i \right)^2
\end{equation}
where $\widehat{\mathcal{Y}}_i$ and $\mathcal{Y}^{*}_i$ are the predicted and the gold standard future values at the position $i$, respectively.
}

\subsection{LLM-driven Time Series Encoding}

{
The LLM processes the temporal representations $\mathbf{Z}^{l-1}$ obtained from the backbone model and the original time series patch sequence $\mathbf{X}$ to compute inherent patterns and semantic relationships among tokens. 
Specifically, both $\mathbf{Z}^{l-1}$ and $\mathbf{X}$ firstly undergo linear projection onto the LLM’s input feature space, forming their embeddings $\mathbf{E}_Z$ and $\mathbf{E}_X$, respectively.
In the input sequence, $\mathbf{E}_Z$ precedes $\mathbf{E}_X$. 
Auxiliary text prompts interleave with these time series embeddings to stimulate the LLM’s reasoning capabilities and guide the learning of semantic connections among time series tokens.
A task-descriptive prompt, such as ``\textit{This is a time series forecasting task. The input contains historical data patterns that need to be analyzed for future predictions.}'', prefixes the entire sequence to frame the forecasting objective.
Immediately before the temporal representations $\mathbf{E}_Z$, the prompt ``\textit{The following are the encoded time series features extracted from the Transformer encoder, which represent the learned temporal patterns.}'' indicates their nature. 
Preceding the original embedding $\mathbf{X}$, the prompt ``\textit{The following are the original patch data features that provide additional context for the prediction task.}'' signals its role. 
These text prompts are tokenized and mapped to their embedding following the standard process, resulting in $\mathbf{P}_{task}$, $\mathbf{P}_{feat}$, and $\mathbf{P}_{data}$, respectively. 
They are then concatenated in the aforementioned order with the time-series features $\mathbf{E}_Z$ and the time-series embedding $\mathbf{E}_X$, forming the LLM input $\mathbf{E}_{LLM}$ by
\begin{equation}
    \mathbf{E}_{LLM} = \text{Cat}(\mathbf{P}_{task}, \mathbf{P}_{feat}, \mathbf{E}_{Z}, \mathbf{P}_{data}, \mathbf{E}_X)    
\end{equation}
Internally, the LLM leverages the knowledge learned from large-scale corpora to uncover intricate patterns within the input features $\mathbf{E}_{LLM}$ and capture high-level semantic relationships.
Following the standard LLM modeling process, we obtain the output from the last layer of the LLM.
We only retain the representations (denoted as $\mathbf{Z}_{LLM}$) corresponding to the positions of the $\mathbf{X}$ and discard all preceding prefixes, including the textual prompts and the temporal representations.
The representation $\mathbf{Z}_{LLM}$ is integrated into the backbone model through the feature fusion module.
}

\subsection{Feature Fusion Module}
{
The feature fusion layer integrates the temporal representation $\mathbf{Z}^{l-1}$ from the ($l$-1)-th layer of the backbone Transformer and the semantic representation $\mathbf{Z}_{LLM}$ from the LLM, obtaining a fused representation $\mathbf{Z}^{l-1}_{E}$ that simultaneously encodes both temporal patterns and semantic information.
Specifically, to fuse $\mathbf{Z}^{l-1}$ and $\mathbf{Z}_{LLM}$, a gate mechanism ($f_{\text{gate}}$) is employed, where the semantic representation $\mathbf{Z}_{LLM}$ and the temporal representation $\mathbf{Z}^{l-1}$ are concatenated and processed through a linear layer followed by a sigmoid activation function, yielding a gate value $g$. 
The unified representation $\mathbf{Z}^{l-1}_{E}$ is then computed by
\begin{equation}
    \mathbf{Z}^{l-1}_{E} = g \cdot \mathbf{Z}_{LLM} + (1 - g) \cdot \mathbf{Z}^{l-1}
\end{equation}
The resulting representation $\mathbf{Z}^{l-1}_{E}$ inherently possesses both temporal dynamics and semantic understanding, and it is subsequently fed into the $l$-th MHA layer in the backbone Transformer model.
}

\section{Experiment Settings}

\subsection{Datasets}

\begin{table}[t]
\centering
\caption{
{The statistics of the datasets, including total timesteps, number of features, and recording period in each dataset.}
}
\vspace{1mm}
\label{tab: dataset}
\begin{tabular}{l|rrr}
\toprule
Dataset & Timesteps & Features  & Period  \\
\midrule
Weather    & \multicolumn{1}{r}{52696}     & \multicolumn{1}{r}{21} & \multicolumn{1}{r}{10 min} \\
ETTh1    & \multicolumn{1}{r}{17420}     & \multicolumn{1}{r}{7} & \multicolumn{1}{r}{1 hour} \\
ETTh2    & \multicolumn{1}{r}{17420}     & \multicolumn{1}{r}{7} & \multicolumn{1}{r}{1 hour} \\
ETTm1    & \multicolumn{1}{r}{69680}     & \multicolumn{1}{r}{7} & \multicolumn{1}{r}{15 min} \\
ETTm2    & \multicolumn{1}{r}{69680}     & \multicolumn{1}{r}{7} & \multicolumn{1}{r}{15 min} \\
ILI    & \multicolumn{1}{r}{966}     & \multicolumn{1}{r}{7} & \multicolumn{1}{r}{1 week} \\
\bottomrule
\end{tabular}%
\end{table}

We run experiments on three widely-used time series datasets: ETT (Electricity Transformer Temperature), Weather, and {Influenza-like Illness (ILI)}, covering diverse domains for comprehensive evaluation.
{The ETT dataset, including ETTh1, ETTh2, ETTm1, ETTm2,} records load and temperature data from electricity transformers, while the Electricity dataset captures hourly power consumption of 321 clients.
The Weather dataset includes 21 meteorological indicators from 1,600 locations, and the ILI dataset tracks weekly influenza-like illness patient ratios. 
All datasets undergo Z-score normalization, where the mean is subtracted and the result is divided by the standard deviation, preserving temporal patterns while mitigating magnitude variations.
Following the protocols in existing studies \cite{autoformer, fedformer}, the data splits maintain temporal order to prevent information leakage: for ETT datasets, sequences are partitioned into 6:2:2 ratios for training, validation, and testing, respectively, whereas other datasets use 7:1:2 splits. 
Such chronological partitioning strategy ensures the evaluation realistically simulates real-world forecasting scenarios where models predict future observations based on historical patterns.
{
The statistics of the datasets (including the number of timesteps, features, and the recording period) are reported in Table \ref{tab: dataset}.
}

\subsection{Baselines and Comparing Models}

{
To validate the effectiveness of our approach, we run a series of baseline models and compare their performance with ours.
Specifically, we implement the following three baselines:
\begin{itemize}[leftmargin=1em]
    \item \textbf{LLM-only}: This baseline directly feeds time series patches into a LLM to obtain corresponding time series representations. These representations are subsequently fed into an linear layer for the subsequent prediction task.
    \item 
    \textbf{Trans-only}: This baseline employs solely the Transformer for encoding the time series. The resulting representation is then used for prediction via an linear projection.
    \item 
    \textbf{Trans-LLM}: This baseline directly adds the features encoded by the backbone Transformer and the features encoded by the LLM without using the feature fusion module.
\end{itemize}
Furthermore, we compare our approach with a series of existing TSF models, which are categorized into the following three groups, namely, Transformer-based models, MLP-based models, and LLM-based models.
For Transformer-based models, we compare our approach with the following ones:
\begin{itemize}[leftmargin=1em]
    \item \textbf{PatchTST}\footnote{{We employ PatchTST/42, which corresponds to the case of a historical window of 336. This aligns with our experimental setup to ensure fair comparison.}} \cite{PatchTST}: This model utilizes a vanilla Transformer to encode time series patches and performs prediction via a linear head.
    \item \textbf{Autoformer} \cite{autoformer}: a Transformer-based model where self-attention is replaced with auto-correlation, combined with a decomposition approach for forecasting.
\end{itemize}
For the MLP-based models, we compare our approach with the following ones:
\begin{itemize}[leftmargin=1em]
    \item \textbf{DLinear} \cite{DLinear}: A single-layer linear prediction model incorporating decomposition approach.
    \item \textbf{Timemixer} \cite{Timemixer}: This model integrates multi-scale information, separately predicts time series components at different scales, and combines them to yield the final forecast.
\end{itemize}
For the MLP-based models, we compare our approach with the following ones:
\begin{itemize}[leftmargin=1em]
    \item \textbf{GPT4TS} \cite{fpt}: This model employs GPT-2 to encode time series patches and conducts prediction through a linear head.
    \item \textbf{Time-LLM} \cite{time-llm}: A model uses text prototypes to reprogram time-series patches. These tokens are then fed into a frozen LLM for encoding, and finally processed via an MLP for forecasting.
\end{itemize}
}

\begin{table*}[th]
\centering
\caption{{Performance of baselines and the proposed model measured by MSE and MAE. "Avg." denotes the aggregated performance across the test sets of eight datasets. For each dataset, the top two results (lower is better) are indicated through \textbf{boldface} (best) and \underline{underlines} (second best), respectively.}
}
\label{tab:baselines}
\begin{tabular}{c|cccccccc}
\toprule
{\multirow{2}{*}{Dataset}} & \multicolumn{2}{c}{\textbf{LLM-only}} & \multicolumn{2}{c}{\textbf{Trans-only}} & \multicolumn{2}{c}{\textbf{Trans-LLM}} & \multicolumn{2}{c}{\textbf{Ours}} \\
{}                           & MSE        & MAE       & MSE          & MAE         & MSE             & MAE      & MSE             & MAE                \\
\toprule
{Weather} & 0.247      & 0.287    & 0.253           & 0.296  & \underline{0.242}        & \underline{0.281}           & \textbf{0.237}          & \textbf{0.279}           \\
{ETTh1} & 0.438      & 0.443   & 0.428           & 0.437   & \underline{0.426}        & \underline{0.434}            & \textbf{0.406}           & \textbf{0.424}        \\
{ETTh2} & 0.403      & 0.422    & \underline{0.352}           & \underline{0.392}   & 0.378        & 0.409           & \textbf{0.333}           & \textbf{0.380}        \\
{ETTm1}  & 0.384      & 0.429   & 0.373           & 0.403    & \underline{0.367}        & \underline{0.395}           & \textbf{0.355}           & \textbf{0.382}        \\
{ETTm2} & 0.337      & 0.341    & \underline{0.270}           & 0.339  & 0.278        & \underline{0.327}            & \textbf{0.258}           & \textbf{0.318}        \\
{ILI} & \underline{1.870}      & \underline{0.896}   & 2.306 & 0.968 & 2.191        & 0.967           & \textbf{1.424}           & \textbf{0.819}        \\
\bottomrule
\end{tabular}%
\end{table*}

\begin{table*}[]
\centering
\caption{
{
Performance of existing models and the proposed model in terms of MSE and MAE. 
}
}
\label{tab:comp_sota}
\resizebox{\linewidth}{!}{
\begin{tabular}{c|cccccccccccccc}
\toprule
{\multirow{2}{*}{Dataset}} & \multicolumn{2}{c}{\textbf{Autoformer}} & \multicolumn{2}{c}{\textbf{PatchTST}} & \multicolumn{2}{c}{\textbf{Timemixer}} & \multicolumn{2}{c}{\textbf{DLinear}} & \multicolumn{2}{c}{\textbf{GPT4TS}}  & \multicolumn{2}{c}{\textbf{Time-LLM}}  & \multicolumn{2}{c}{\textbf{Ours}}\\
{}                           & MSE        & MAE       & MSE          & MAE         & MSE             & MAE      & MSE             & MAE   & MSE             & MAE       & MSE             & MAE  & MSE             & MAE    \\
\toprule
{Weather} & 0.338      & 0.382     & \underline{0.225}        & \underline{0.264}             & 0.240     & 0.271           & 0.249    & 0.300           & 0.237   & 0.270           & \textbf{0.225}   & \textbf{0.257}           & 0.237   & 0.279\\
{ETTh1} & 0.496      & 0.487     & 0.413        & 0.430             & 0.447      & 0.440           & 0.423    & 0.437           & 0.427   & 0.426           & \underline{0.408}  & \underline{0.423}           & \textbf{0.406} & \textbf{0.422}\\
{ETTh2} & 0.450      & 0.459     & \textbf{0.330}        & \textbf{0.379}            & 0.364      & 0.395           & 0.431    & 0.447           & 0.346  & 0.394           & 0.334   & \underline{0.383}           & \underline{0.333} & \textbf{0.379}\\
{ETTm1}  & 0.588      & 0.517     & \underline{0.351}        & 0.387            & 0.381      & 0.395           & 0.357   & \underline{0.378}           & 0.352  & 0.383           & \textbf{0.329}   & \textbf{0.372}           & 0.355  & 0.382\\
{ETTm2} & 0.327      & 0.371     & \underline{0.255}        & \underline{0.315}           & 0.275      & 0.323           & 0.267  & 0.334           & 0.266   & 0.326           & \textbf{0.251}  & \textbf{0.313}           & 0.258  & 0.318\\
{ILI} & 3.006      & 1.161     & 1.443        & \textbf{0.798}             & 2.037      & 0.938           & 2.169   & 1.041           & 1.925   & 0.903           & \underline{1.435}  & \underline{0.801}           & \textbf{1.424}  & 0.819 \\
\bottomrule
\end{tabular}%
}
\end{table*}

\subsection{Implementation Details}
{
For time series length configurations, the historical sequence length is fixed at 104 for the ILI dataset, while other datasets adopt a default historical length of 336. 
The forecast horizon is evaluated across four settings $P \in \{96, 192, 336, 720\}$, which comprehensively assesses model performance for short-term, medium-term, and long-term forecasting tasks. This setup ensures consistency with prior benchmarks while testing generalization capabilities under varying temporal dependencies.
For patch splitting, following the setting from \cite{PatchTST}, we use a patch length of 24 and a stride of 2 on the ILI dataset, while employing a patch length of 16 and a stride of 8 uniformly across all other datasets.
We employ a channel separation technique for processing multivariate time series.
Additionally, the time series undergoes reversible instance normalization (RevIN), whereby the mean and variance are computed from the historical time series prior to model input and used for normalization
The corresponding statistics are then added back to the model’s predicted outputs. This technique has been demonstrated to effectively address the distribution-shift problem \cite{reversibleIN}.
}

{
Regarding the model architecture, we employ a time series Transformer encoder based on \citet{PatchTST}, which utilizes a 3-layer transformer encoder, where the LLM output is used to enhance the output of the second layer.
On the ILI dataset, the model dimension is set to 128 with 16 attention heads, while a uniform configuration of a model dimension of 16 and 4 attention heads is used for all other datasets. 
For the LLM, we adopt GPT-2 \cite{gpt2} (with 768-dimensional hidden vectors and 2 layers) and apply Low-Rank Adaptation (LoRA) \cite{lora} for parameter-efficient fine-tuning. The overall model is trained for 300 epochs with a learning rate of 0.0001.
}

{
Following existing studies \cite{autoformer, PatchTST}, we use Mean Squared Error (MSE) and Mean Absolute Error (MAE) as the primary evaluation metrics to quantify forecasting accuracy. 
Both metrics are computed across all prediction horizons, with lower values indicating better performance. To ensure statistical reliability, all reported results represent averaged outcomes from three independent training runs with different random seeds.
}

\section{Results and Analyses}

\subsection{Overall Results}

{
To validate the effectiveness of the proposed approach, we compare the performance of our approach with the baselines and report the results in Table \ref{tab:baselines}.
Several observations emerge from the results.
First, Transformer-only baseline exhibits lower error values than LLM-only baseline on ETT datasets but higher errors on other datasets.
This indicates different performance characteristics for these two models across different datasets.
This occurs because the representations learned by the LLM-only model possess stronger semantic patterns, whereas the Transformer-only model learns representations capturing stronger temporal dynamics. These distinct representations contribute differently to forecasting performance depending on the dataset.
Second, the Transformer-LLM model consistently achieves lower error values than both the Transformer-only and LLM-only baselines.
This demonstrates that combining the two models provides superior time series representations. 
This improvement arises because the Transformer and LLM learn complementary characteristics of the time series, capturing different aspects and levels of information.
Integrating these representations enhances prediction accuracy. 
Finally, our proposed model attains the smallest prediction MSE and MAE errors.
This signifies that adaptively fusing the two types of representations better leverages their respective strengths. 
The rationale is that the representations learned by the two models exhibit heterogeneity.
Therefore, directly combining the two representations leads to confusion, preventing the predictor from effectively utilizing useful information for TSF.
These findings confirm the effectiveness of our proposed approach.
}

{
The comparative results between our approach and existing time series forecasting approaches are reported in Table \ref{tab:comp_sota}, yielding the following observations.
First, the proposed approach achieves the lowest average MSE and average MAE error values, demonstrating its superior forecasting accuracy.
Particularly, our approach outperforms Transformer-based and MLP-based models, achieving lower errors on ETTh1 dataset. 
This indicates that integrating LLMs as a complement to conventional deep learning models enhances prediction performance. 
The key factor is the LLM's reasoning capability acquired through corpus pre-training, enabling it to learn high-level semantic patterns from time series data. 
This capability effectively offsets the limitations of conventional models in capturing such abstract information. 
Finally, compared to other LLM-based models, our approach achieves lower error rates on ETTh1 and ETTh2 datasets. 
This advantage stems from our utilization of the LLM: instead of directly using it in time series prediction, we employ the LLM to generate semantic features, which are then provided to deep learning models that specialize in processing continuous numerical data. 
This strategy circumvents the issue of natural language bias in LLM representations, which may hurt prediction accuracy when utilized directly by predictors lacking corresponding linguistic knowledge.
}

\subsection{Effects of LLM Encoding}

\begin{figure}[t]
    \centering
    \includegraphics[width=\linewidth]{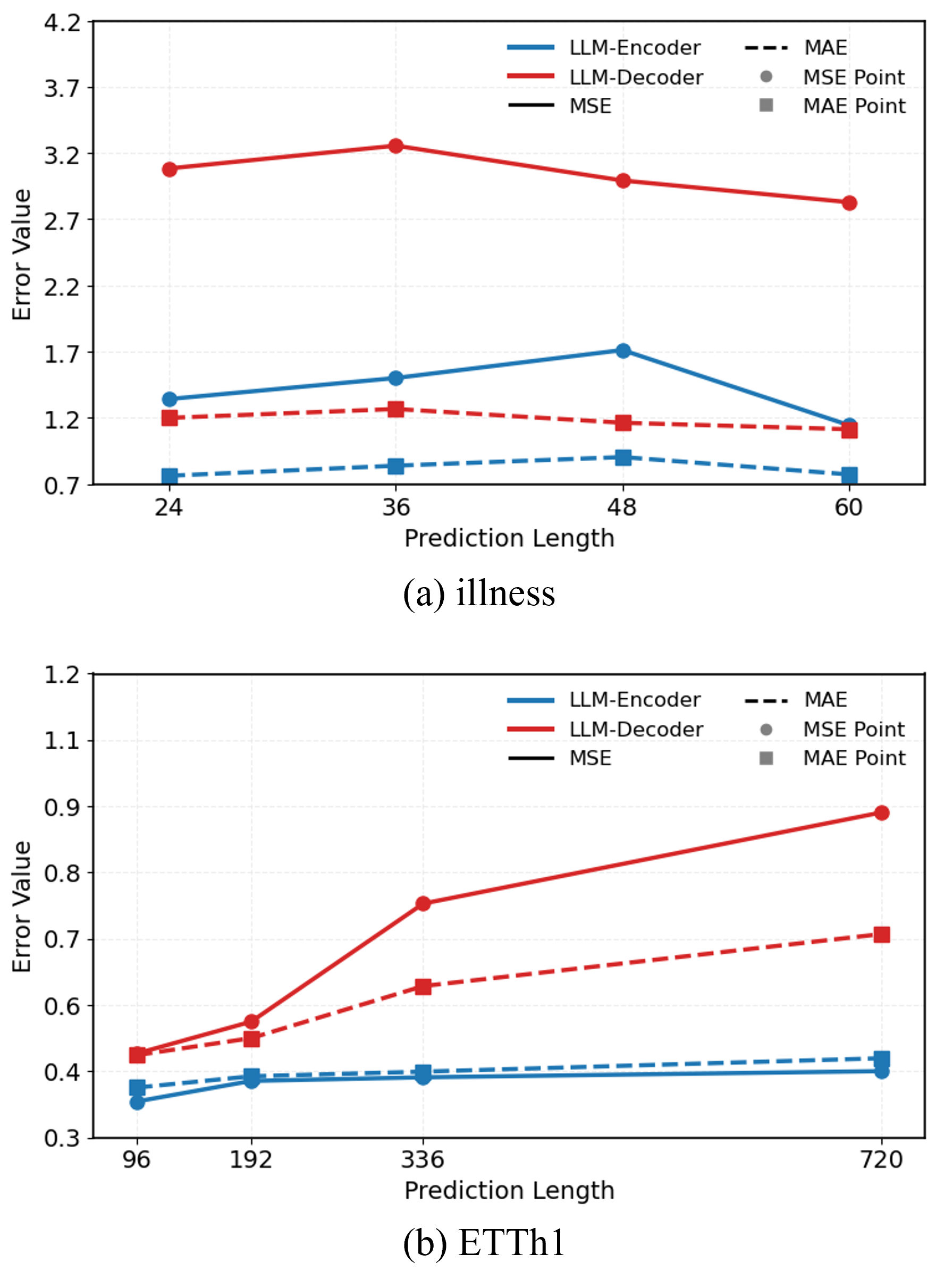}
    \vspace{-1cm}
    \caption{{Comparison of our approach and the model that directly uses the LLM in our approach to perform TSF in a non-autoregressive manner on the ILI and ETTh1 datasets.}}  
    \label{fig: llm_encoding}
\end{figure}

{
In our setup, a decoder-only LLM serves as the time series encoder, with its generated representations used for subsequent processing and prediction. 
To investigate the role of the LLM specifically as a time series encoder, we compare its performance when employed as a time series "decoder". 
Specifically, we modify the proposed model to achieve a non-autoregressive LLM decoding approach: append placeholder tokens matching the number of patches corresponding to the prediction length to the end of the input token sequence, then remove the subsequent feature fusion layer and prediction head.
These tokens serve as placeholders for the patches to be predicted; the LLM predicts the values at these positions in a single forward pass.
Finally, we extract the representations only from these placeholder tokens and apply a linear layer to invert the patching operation, reconstructing the original sequence shape. 
Figure \ref{fig: llm_encoding} presents the results of this model and our full model on the ILI and ETTh1 datasets across different prediction lengths.
First, we observe a higher prediction error for the LLM-decoder compared to our proposed model on the ILI dataset.
This indicates that the LLM-decoder model fails to capture useful information for the task from this specific data.
Furthermore, the LLM-decoder obtains comparable error metrics to our proposed model on the ETTh1 dataset for the short-term forecasting horizon of 96 steps.
This result suggests that employing an LLM for decoding delivers acceptable short-term forecasting performance on the ETT dataset, primarily because temporal dynamics show relatively consistent and stable characteristics within short time windows. 
However, the performance of LLM decoding prediction degrades as the prediction length increases.
This degradation demonstrates the LLM failure to correctly decode continuous variation patterns in historical time series, causing progressive deviation from ground truth values during long-term forecasting.
}

\subsection{Effects of Time Series Representation}
\begin{figure*}[t]
    \centering
    \includegraphics[width=\linewidth]{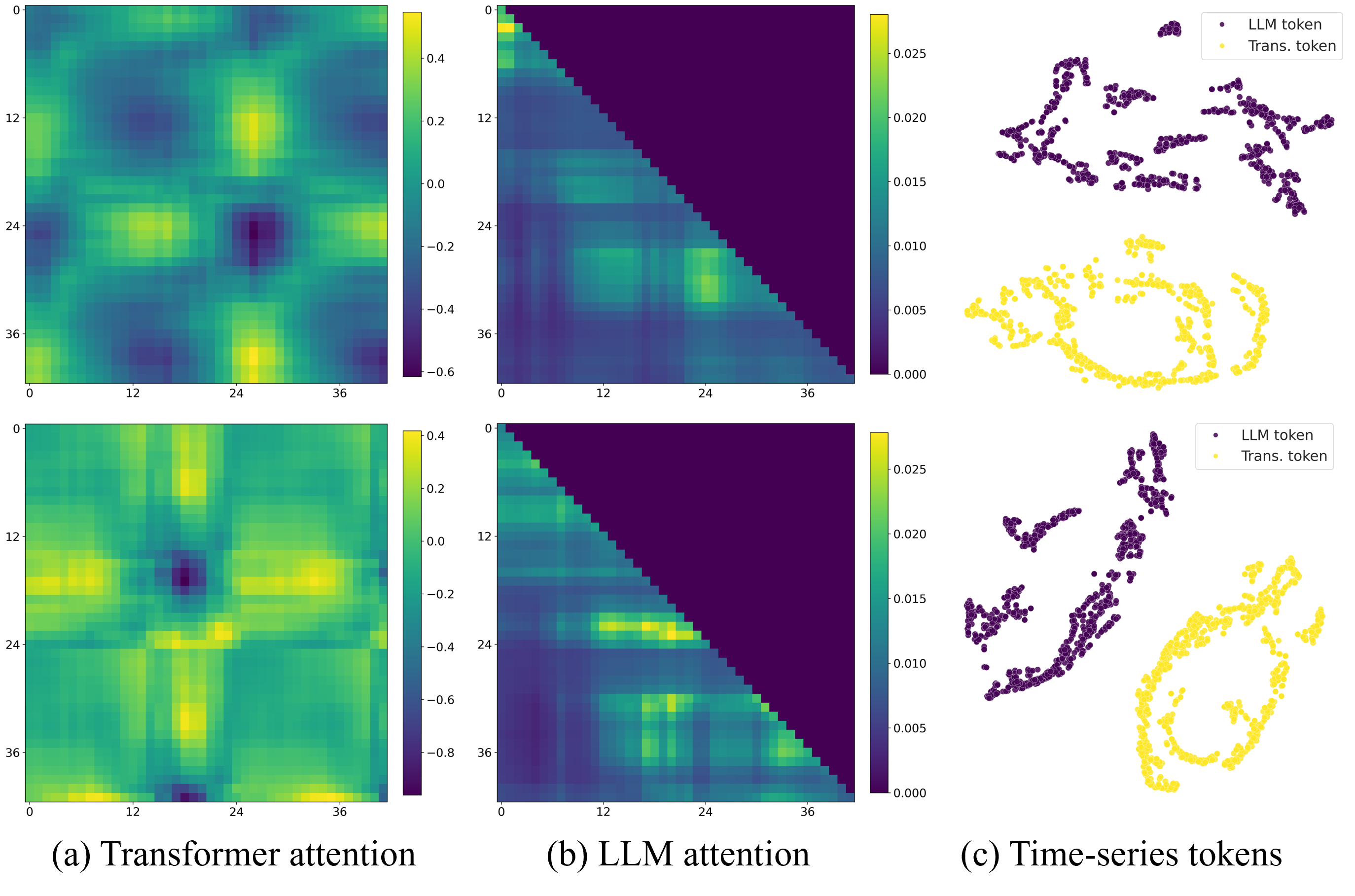}
    \vspace{-1cm}
    \caption{{(a) Attention visualization from the last layer of our approach's Transformer encoder and (b) the second layer of the LLM. (c) t-SNE visualization of time-series tokens from the Transformer encoder and the LLM.}}
    \label{fig: attne & t-sne}
\end{figure*}
\begin{figure*}[t]
    \centering
    \includegraphics[width=\linewidth]{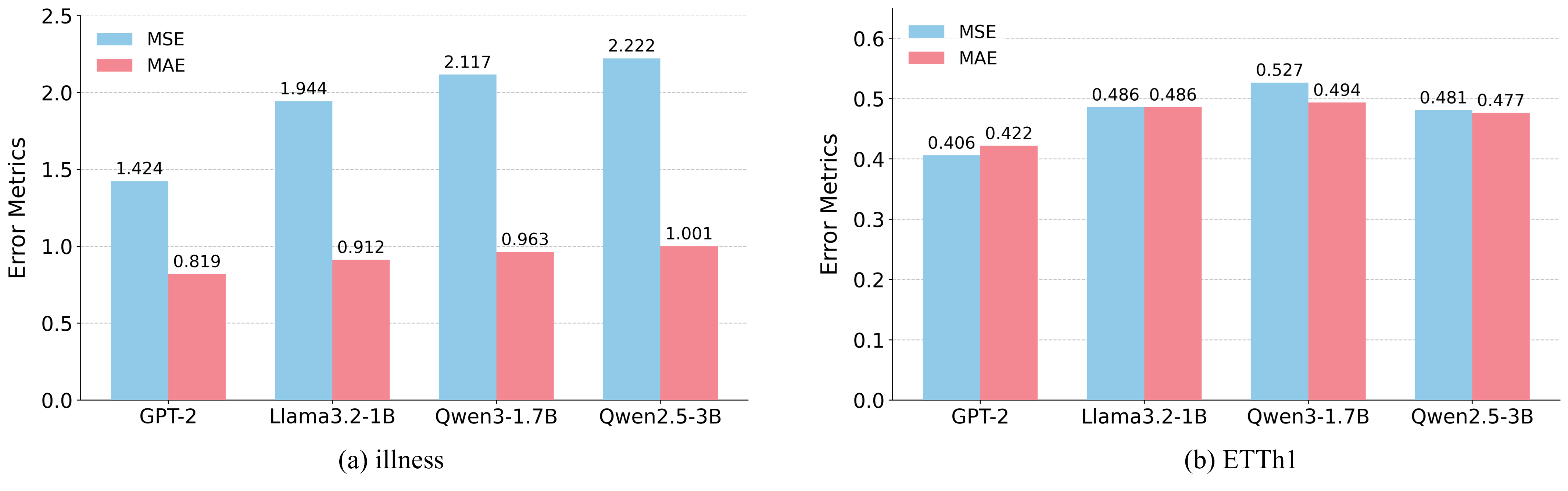}
    \vspace{-1cm}
    \caption{{Comparison of different types and sizes of LLMs as time series encoders for forecasting. We report MSE and MAE for each LLM encoder on the ETTh1 dataset.}}
    \label{fig:llm type}
\end{figure*}
\begin{figure}[t]
    \centering
    \includegraphics[width=\linewidth]{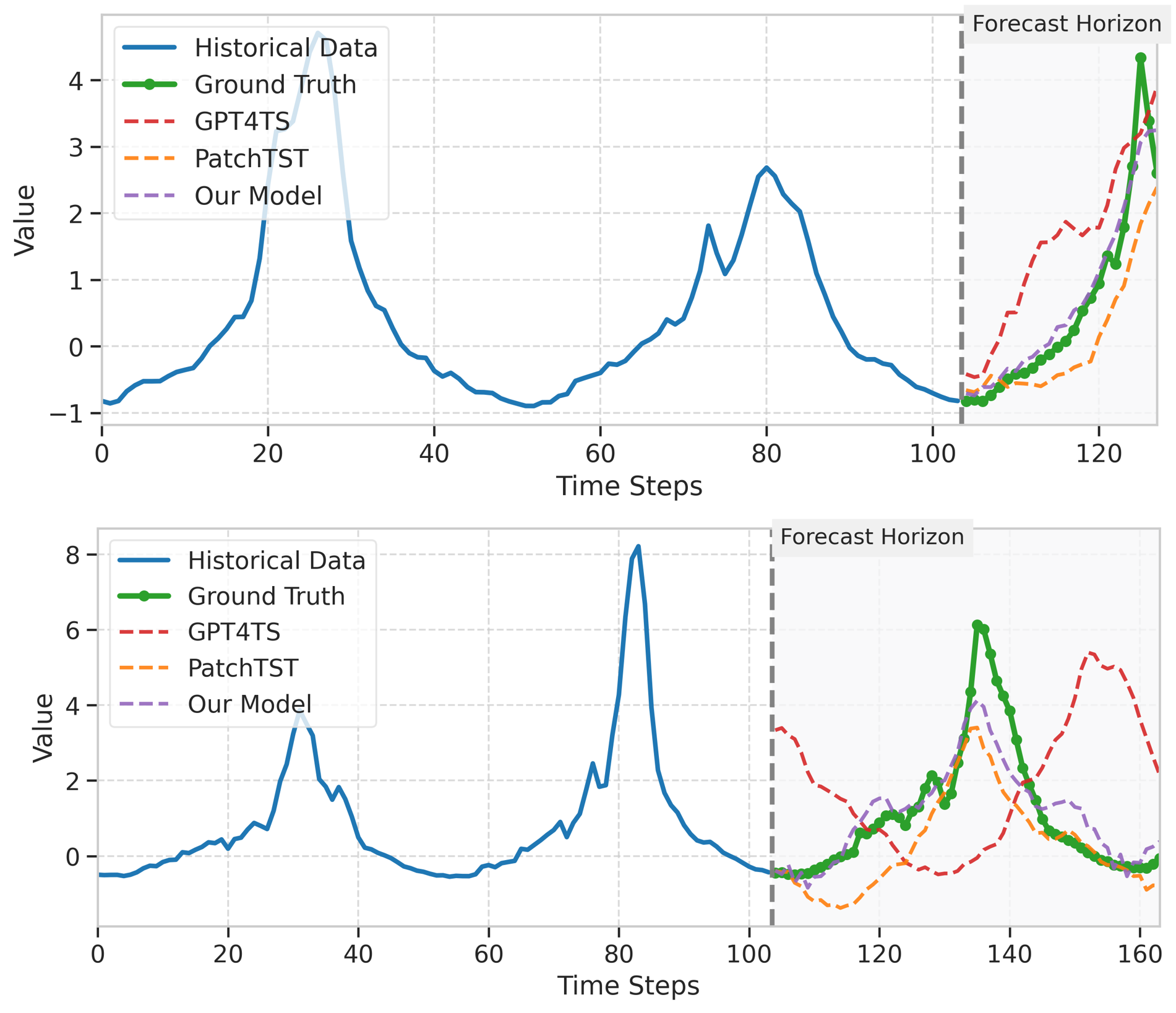}
    \vspace{-1cm}
    \caption{{Case study on the ETTh1 dataset. The historical time series is plotted in blue. The future time series to be forecasted is indicated by a dashed gray line. Forecasts from the PatchTST, GPT4TS, and our proposed model are depicted in yellow, green, and red colors, respectively.}}
    \label{fig: case study}
\end{figure}

{
To investigate the differences between representations learned by the Transformer and those learned by the LLM for time series, we visualize attention weights for time series patches from the final Transformer encoder block and the final LLM layer. 
We also display t-SNE plots of the time series tokens encoded by the Transformer and the LLM before feature fusion.
The heatmaps are shown in Figure \ref{fig: attne & t-sne}.
Several observations emerge.
First, the attention maps in the Transformer show a continuous block-style attention distribution, indicating that the learned relational patterns tend to present continuous temporal variations.
In contrast, the attention heatmaps in the LLM display a sparser attention distribution, suggesting the LLM places greater emphasis on analyzing and extracting critical patches, which is similar to extracting keywords from a sentence for semantic summarization.
This attention distribution difference reflects the distinct intrinsic learning mechanisms of the two models.
The t-SNE visualization reveals that tokens encoded by the Transformer and tokens encoded by the LLM form distinct clusters. 
This separation indicates that the Transformer-encoded tokens and LLM-encoded tokens possess different distribution characteristics.
This divergence originates from their differing internal learning processes, leading the Transformer to encode more information about temporal dynamics within its tokens, while the LLM encodes more information about semantic patterns.
These findings demonstrate the source of performance gains from integrating these complementary information types for prediction.
}

\subsection{Effects of LLM Type and Scale}
{
To investigate the impact of employing diverse LLM types and sizes as time series encoders, we test LLMs including GPT-2, Llama3.2-1B \cite{llama3}, Qwen3-1.7B \cite{qwen3}, and Qwen2.5-3B \cite{qwen25}. 
The results of these models are presented in Figure \ref{fig:llm type}. 
The results first reveal that when replacing GPT-2 with Llama, the error metric increases. Transitioning from Llama to Qwen leads to a further rise in error.
Concurrently, the model size progressively grows larger throughout this sequence of LLM substitutions.
These findings demonstrate that more advanced and larger-scale LLMs do not yield improved forecasting performance. 
This occurs because, in our framework, the LLM only serves as an encoder for time series data, extracting semantic representations to help TSF.
Since LLMs inherently rely on their natural language priors to handle unseen tokens, an overly strong linguistic bias hinders the model's ability to rapidly learn the correct processing approach from limited time series data. 
These results indicate that utilizing excessively large LLMs for time series semantic representation is unnecessary. Such large models increase training costs and cause performance degradation.
}

\subsection{Case Study}
{
To qualitatively investigate the impact of the proposed approach on TSF, Figure \ref{fig: case study} presents prediction results from GPT4TS, PatchTST, and our proposed model on the ILI dataset.
Several observations emerge from the results.
First, the short-term prediction scenario in the upper plot shows our approach consistently generates curves closer to the ground truth than other approaches.
This indicates our model accurately captures continuous variation patterns within the time series, enabling high-precision short-term forecasts. 
This occurs because the Transformer encoder within our model effectively extracts temporal dynamics from historical data, enabling the prediction of similar sequential variations.
Second, the long-term prediction scenario in the lower plot reveals that GPT4TS produces predictions opposite to the ground truth.
While PatchTST correctly estimates the future trend direction, its predictions lack high precision, showing a significant gap from the ground truth curve.
Our proposed model, however, not only correctly identifies the trend but also produces a prediction curve closely aligned with the ground truth, demonstrating high forecasting accuracy. 
This demonstrates the robustness of our proposed approach relative to other models.
The LLM component learns higher-level semantic relationships within the time series, preventing over-reliance on simple, repetitive patterns.
These findings collectively validate the effectiveness of the proposed approach.
}

\section{Related Work}

\subsection{Time Series Forecasting}
{
Early time series forecasting models use statistical approaches, such as the autoregressive integrated moving average model (ARIMA) \cite{arima1, arima2} and exponential smoothing (ETS) \cite{ets-tsf}.
These models represent time series through linear parametric equations and handle trends and seasonality via differencing or decomposition.
These models struggle with complex nonlinear patterns, long-range dependencies, and high-dimensional data in time series, which necessitates significant manual tuning and limits scalability. 
Therefore, research shifts towards automated feature learning and deep learning models that offer higher flexibility. 
Early deep learning approaches employ recurrent neural networks (RNNs) \cite{rnn-rw1, rnn-rw2} and long short-term memory networks (LSTMs) \cite{lstm-rw} to capture temporal dependencies.
Subsequently, convolutional neural networks (CNNs) \cite{TSConv2, cnn_tsf_rw} are utilized to efficiently extract local patterns using dilated convolutions. 
More recently, researchers apply Transformer architectures \cite{autoformer, fedformer} to TSF, leveraging self-attention to model global dependencies and interactions across arbitrary time lags effectively.
Current research focuses on large foundation time series models.
For instance, \citet{timer} build dedicated time series foundation models through massive pretraining. 
Simultaneously, multimodal forecasting gains traction \cite{time-llm, chattime, mcd-tsf}, merging time series with external data (such as text, images, graphs, etc.) for richer context-aware TSF and thus improving the model performance.
Our approach focuses on how to apply natural language knowledge from LLMs to time series forecasting.
}

\subsection{LLM-based Time Series Modeling}

{
LLMs demonstrate remarkable capabilities across diverse natural language processing (NLP) tasks \cite{llm_nlp1,llm_nlp2,llama3,llm-nlp3}, achieving exceptional performance on numerous benchmarks.
This success drives a trend of adapting these powerful foundation models to analyze diverse data modalities beyond text \cite{llm_cv, llm_mm}, including time series data.
Therefore, researchers increasingly explore leveraging LLMs for time series forecasting. 
Early approaches \cite{llmtime, TEST} treat time series data as textual sequences, directly feeding flattened numerical values or simple symbolic representations into the LLM for prediction tasks.
Subsequent approaches \cite{fpt, tempo, fpt++} advance by embedding raw time series data into continuous vectors compatible with LLM input layers, enabling the model to process numerical sequences more effectively.
To better leverage the rich world knowledge learned by the LLMs, more recent efforts focus on aligning the time series modality with the textual modality \cite{timecma, calf} so as to translate temporal patterns into concepts the LLM understands.
These models directly apply the LLM to modeling the time series data and show satisfying performance. 
However, standard LLMs are initially designed for discrete token sequences with long-range dependencies, and thus struggle to capture the continuous temporal dynamics inherent in time series data.
Their architectures lack inductive biases specifically designed for such temporal structures.
Differing from these studies, we propose leveraging the LLM's capability to learn semantic relationships by using it as a feature extractor. 
We then integrate these LLM-derived semantic features as supplementary information alongside conventional numerical time series features within a downstream model specifically designed for TSF.
This integration aims to enhance forecasting performance through enriched contextual understanding.
}

\section{Conclusion}

{
In this paper, we present a novel approach integrating the reasoning capabilities of LLMs with conventional time-series Transformer forecasting models. 
Our approach utilizes the raw time-series data along with its Transformer-encoded temporal features as input to the LLM. 
Leveraging the LLM's capability for discrete token reasoning and pattern learning, the model actively captures high-level semantic patterns from the time-series data. 
An adaptive feature fusion layer dynamically integrates these extracted semantic representations with the original Transformer-encoded temporal features.
A linear predictor then utilizes this fused representation for final time-series forecasting. 
Comprehensive experiments on diverse real-world datasets demonstrate the effectiveness and superior performance of our approach.
Future research directions include further incorporating textual features for enhanced fusion and exploring multi-modal collaborative forecasting strategies.
}

\bibliographystyle{acl_natbib}
\bibliography{my_paper}
\iftaclpubformat

\onecolumn

\appendix

\end{document}